\documentclass[10pt,twocolumn,letterpaper]{article}

\usepackage[pagenumbers]{cvpr} %

\usepackage{bm}
\usepackage{tabularx}
\usepackage{booktabs}
\usepackage{multirow}
\usepackage{tikz}
\usepackage{twemojis}

\newcolumntype{Y}{>{\centering\arraybackslash}X}

\definecolor{c_red}{HTML}{ea4335}
\definecolor{c_green}{HTML}{34a853}
\colorlet{tablered}{c_red!99!black}
\colorlet{tablegreen}{c_green!99!black}

\newcommand{\PAR}[1]{\vskip4pt \noindent {\bf #1~}}
\newcommand{\PARbegin}[1]{\vskip1pt \noindent {\bf #1~}}

\newcommand{\tablestep}[1]{\textcolor{gray!90}{(#1)}}

\newcommand{\downrightarrow}{\hspace{0pt}%
  \raisebox{1.5pt}{\begin{tikzpicture}[scale=0.4, baseline=(current bounding box.south)]
    \draw[-latex] (0.0, 0.5) -- (0.0, 0.25) -- (0.55, 0.25);
  \end{tikzpicture}} %
}

\newcommand{\ours}{Plain Mask Transformer}
\newcommand{\oursShort}{PMT}
\newcommand{\decoder}{Plain Mask Decoder}
\newcommand{\decoderShort}{PMD}

\makeatletter
\newcommand\notsotiny{\@setfontsize\notsotiny{6.5}{7.5}}
\newcommand\notscriptsize{\@setfontsize\notscriptsize{7.5}{8.5}}
\makeatother

\usepackage{pgfplots}
\pgfplotsset{compat=1.18}
\usetikzlibrary{arrows.meta, backgrounds, calc}
\usepackage{fontawesome5}

\expandafter\def\expandafter\normalsize\expandafter{%
    \normalsize%
    \setlength\abovedisplayskip{2pt}%
    \setlength\belowdisplayskip{2pt}%
    \setlength\abovedisplayshortskip{2pt}%
    \setlength\belowdisplayshortskip{2pt}%
}

\definecolor{cvprblue}{rgb}{0.21,0.49,0.74}
\usepackage[pagebackref,breaklinks,colorlinks,allcolors=cvprblue]{hyperref}

\def\paperID{27} %
\def\confName{CVPR}
\def\confYear{2026}

\title{PMT: Plain Mask Transformer for Image and Video Segmentation \\ with Frozen Vision Encoders}

\author{Niccol\`{o} Cavagnero \quad Narges Norouzi \quad Gijs Dubbelman \quad Daan de Geus\\[0.6em]
Eindhoven University of Technology
}

\begin{document}
\maketitle
\begin{abstract}
Vision Foundation Models (VFMs) pre-trained at scale enable a single frozen encoder to serve multiple downstream tasks simultaneously.
Recent VFM-based encoder-only models for image and video segmentation, such as EoMT and VidEoMT, achieve competitive accuracy with remarkably low latency, yet they require finetuning the encoder, sacrificing the multi-task encoder sharing that makes VFMs practically attractive for large-scale deployment.
To reconcile encoder-only simplicity and speed with frozen VFM features, we propose the \decoder{} (\decoderShort{}), a fast Transformer-based segmentation decoder that operates on top of frozen VFM features.
The resulting model, the \ours{} (\oursShort{}), preserves the architectural simplicity and low latency of encoder-only designs while keeping the encoder representation unchanged and shareable.
The design seamlessly applies to both image and video segmentation, inheriting the generality of the encoder-only framework.
On standard image segmentation benchmarks, \oursShort{} matches the frozen-encoder state of the art while running up to ${\sim}3\times$ faster. For video segmentation, it even performs on par with fully finetuned methods, while being up to $8\times$ faster than state-of-the-art frozen-encoder models. Code: \href{https://github.com/tue-mps/pmt}{https://github.com/tue-mps/pmt}.
\end{abstract}
    
\section{Introduction}
\label{sec:intro}

Vision Foundation Models (VFMs), pre-trained at scale on large and diverse datasets, have established the Vision Transformer (ViT)~\cite{dosovitskiy2021vit} as the dominant encoder in modern computer vision.
The DINO family~\cite{caron2021dino, oquab2023dinov2, simeoni2025dinov3} exemplifies this trend: by combining the ViT architecture with self-supervised objectives that promote dense and semantically rich representations, these models achieve strong performance in a broad range of downstream tasks without any task-specific design choices in the encoder itself.

\definecolor{adapterlinecolor}{RGB}{31, 119, 180} %
\definecolor{eomtcolor}{RGB}{255, 127, 14}        %

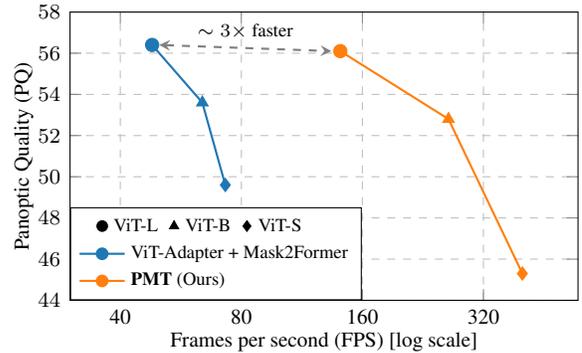
\begin{figure}[t]
    \centering
    \begin{tikzpicture}[tight background]        
        \def\vitLmark{*}
        \def\vitBmark{triangle*}
        \def\vitSmark{diamond*}

        \def\vitadapterLdata{(48, 56.4)}
        \def\vitadapterBdata{(64, 53.6)}
        \def\vitadapterSdata{(73, 49.6)}

        \def\pmtLdata{(141, 56.1)}
        \def\pmtBdata{(262, 52.8)}
        \def\pmtSdata{(400, 45.3)}

        \begin{axis}[
            font=\small,
            width=1.0\linewidth,
            height=0.65\linewidth,
            xmode=log,
            xlabel={Frames per second (FPS) [log scale]},
            ylabel={Panoptic Quality (PQ)},
            ylabel style={yshift=-0.5em},
            xlabel style={yshift=0.4em},
            xmin=30, xmax=550,
            ymin=44, ymax=58,
            xtick={40, 80, 160, 320},
            xticklabels={40, 80, 160, 320},
            ytick={44, 46, 48, 50, 52, 54, 56, 58},
            legend style={font=\scriptsize, at={(0.0,0.0)}, anchor=south west},
            legend cell align=left,
            ymajorgrids=true,
            xmajorgrids=true,
            grid style=dashed,
            tick label style={font=\footnotesize},
            label style={font=\footnotesize},
        ]

        \addlegendimage{empty legend}
        \addlegendentry{\hspace{-11.7pt}%
        \tikz \node[mark size=2pt, inner sep=0.8pt] at (0, 0) {\pgfuseplotmark{\vitLmark}};~~ViT-L~~
        \tikz \node[mark size=2pt, inner sep=0.8pt] at (0, 0) {\pgfuseplotmark{\vitBmark}};~~ViT-B~~
        \tikz \node[mark size=2pt, inner sep=0.8pt] at (0, 0) {\pgfuseplotmark{\vitSmark}};~~ViT-S
        };

        \addlegendimage{color=adapterlinecolor, mark=*, line width=0.8pt}
        \addlegendentry{ViT-Adapter + Mask2Former}
        
        \addlegendimage{color=eomtcolor, mark=*, line width=0.8pt}
        \addlegendentry{\textbf{PMT} (Ours)}

        \draw[<->, >=stealth, thick, gray, dashed] 
            (axis cs:50, 56.4) -- (axis cs:134, 56.1)
            node[midway, above, font=\scriptsize, text=black] {$\sim 3 \times$ faster};

        \addplot[color=adapterlinecolor, line width=0.8pt, mark=none] coordinates {\vitadapterLdata \vitadapterBdata \vitadapterSdata};
        \addplot[only marks, mark=\vitLmark, mark size=2.5pt, color=adapterlinecolor] coordinates {\vitadapterLdata};
        \addplot[only marks, mark=\vitBmark, mark size=2.5pt, color=adapterlinecolor] coordinates {\vitadapterBdata};
        \addplot[only marks, mark=\vitSmark, mark size=2.5pt, color=adapterlinecolor] coordinates {\vitadapterSdata};

        \addplot[color=eomtcolor, line width=0.8pt, mark=none] coordinates {\pmtLdata \pmtBdata \pmtSdata};
        \addplot[only marks, mark=\vitLmark, mark size=2.5pt, color=eomtcolor] coordinates {\pmtLdata};
        \addplot[only marks, mark=\vitBmark, mark size=2.5pt, color=eomtcolor] coordinates {\pmtBdata};
        \addplot[only marks, mark=\vitSmark, mark size=2.5pt, color=eomtcolor] coordinates {\pmtSdata};
        
        \end{axis}
    \end{tikzpicture} 
    \vspace{-5pt}
    \caption{\textbf{ViT-Adapter + Mask2Former \vs \oursShort{} (Ours).} \oursShort{} exhibits a better trade-off between Panoptic Quality and FPS across different sizes of \textit{frozen} DINOv3~\cite{simeoni2025dinov3} pre-trained ViTs~\cite{dosovitskiy2021vit}. Evaluated on COCO \textit{val2017}~\cite{lin2014coco}, see \cref{tab:model_size}.}
    \label{fig:comparison}
\end{figure}

The representational richness of VFMs enables a radical rethinking of downstream architectures.
For instance, consider the task of image segmentation, which requires an image to be divided into pixel-level masks.
Until recently, state-of-the-art ViT-based image segmentation models applied various task-specific components on top of the ViT, such as a convolutional adapter~\cite{chen2023vitadapter}, a pixel decoder, and a Transformer decoder~\cite{cheng2021maskformer, cheng2022mask2former, jain2023oneformer}.

While this approach is effective, the Encoder-only Mask Transformer (EoMT)~\cite{kerssies2025eomt} method showed that these task-specific components are, in fact, largely unnecessary in combination with current VFMs.
Specifically, EoMT consists of a plain ViT whose final layers are injected with a set of learnable queries, which are then processed alongside the patch tokens.
During this process, each query accumulates the information needed to predict a binary segmentation mask and class label, allowing it to yield segmentation predictions with competitive accuracy at a fraction of the latency.
The \textit{encoder-only} framework was recently extended to the video domain, with VidEoMT~\cite{norouzi2026videomt}. 
Following the principles of EoMT, this model replaces task-specific tracking modules and temporal Transformer layers with a lightweight query propagation mechanism alongside the ViT, achieving $5$--$10\times$ faster inference over prior art.
Crucially, the success of this encoder-only paradigm requires sufficient scale, both in pre-training data and model size: without a sufficiently large and well-trained encoder, removing the specialized modules leads to a significant accuracy drop~\cite{kerssies2025eomt, norouzi2026videomt}.
Only at scale does the pre-trained encoder alone carry the representational capacity previously distributed across many specialized modules, making those components largely redundant.

However, both EoMT and VidEoMT require \emph{finetuning the entire encoder}.
On the one hand, patch token representations must be updated to accommodate the segmentation task, requiring full finetuning of the encoder.
More critically, because queries are injected into the self-attention layers, the pre-trained weights must adapt to incorporate these new tokens. As we empirically verify, freezing the encoder is not merely suboptimal, it fundamentally prevents the mechanism from working, as the pre-trained attention layers have no notion of the injected queries.

The fact that EoMT and VidEoMT require finetuning the ViT presents practical limitations.
Specifically, by updating the encoder parameters for a particular segmentation task and dataset, the VFM can no longer be used for any other downstream task. Each task and dataset require a separate ViT with finetuned parameters.
As a consequence, if predictions are needed for multiple tasks or class definitions during deployment, one must either (a) maintain separate finetuned encoder-only models for each task, or (b) use a single frozen VFM encoder with inefficient task-specific decoders on top~\cite{simeoni2025dinov3}.
Clearly, both options are inefficient.

Therefore, in this work, we apply the philosophy pioneered by EoMT to the setting where the VFM is kept frozen.
Specifically, given a frozen ViT encoder, we explore how much the task-specific decoders for image and video segmentation can be simplified when the ViT encoder is sufficiently large and pre-trained with large-scale data.
While most task-specific modules could be removed without accuracy drop in the finetuned-encoder setting~\cite{kerssies2025eomt}, we observe more substantial drops in combination with a frozen encoder.
To compensate for this loss in accuracy without introducing substantial computational overhead, we present \ours{} (\oursShort{}). 
\oursShort{} mimics the behavior of the last $L_2$ encoder layers of EoMT and VidEoMT with a small Transformer decoder, called the \decoder{} (\decoderShort{}). This \decoderShort{} takes the learnable queries and features from the frozen ViT encoder, and processes them jointly through regular Transformer layers, just like the last $L_2$ layers of EoMT.
As this is a very lightweight module that only uses standard Transformer operations, this design preserves the architectural simplicity  and low latency of encoder-only methods, while the encoder remains frozen and shareable across multiple downstream tasks.
Importantly, this design seamlessly applies to both image and video segmentation, inheriting the generality of the encoder-only framework.

Through experiments, we find that \oursShort{} matches the accuracy of state-of-the-art frozen-encoder models for image segmentation while being up to $3\times$ faster. Moreover, for video segmentation, it can even compete or outperform \textit{fully finetuned} state-of-the-art methods, while being up to $8\times$ faster than frozen-encoder baselines.
Together, these results demonstrate the effectiveness of \oursShort{} for image and video segmentation.
Notably, \oursShort{} obtains these results while keeping the encoder frozen, making it directly compatible with multi-task deployment where a single shared encoder must serve multiple tasks simultaneously.

In summary, we make the following contributions.
\begin{itemize}
    \item We identify and experimentally verify a fundamental limitation of encoder-only segmentation models: their approach of injecting learnable queries into the ViT encoder is not compatible with keeping the encoder frozen.
    \item To overcome this limitation, we propose the \decoder{} (\decoderShort{}), a lightweight decoder that operates on frozen VFM features, reconciling the architectural simplicity and speed of encoder-only methods with the frozen-encoder paradigm. We term the resulting approach the \ours{} (\oursShort{}).
    \item We find that \oursShort{} matches state-of-the-art frozen-encoder models on image segmentation while being up to $3\times$ faster, and that it performs on par with fully finetuned methods for video segmentation at speeds up to $8\times$ higher than frozen-encoder baselines.
\end{itemize}

\section{Related Work}
\label{sec:related}

\PARbegin{Image Segmentation.}
Image segmentation is a fundamental computer vision task that requires dividing an image into pixel-level segments, each associated with a class label.
Traditionally, segmentation methods relied on per-pixel classification~\cite{long2015fcn, chen2018deeplab, chen2018deeplabv3p}, assigning a label to each pixel.

Recently, the Mask Transformer paradigm~\cite{carion2020detr, cheng2021maskformer, wang2021maxdeeplab} has enabled a unified mask-classification formulation for semantic, instance, and panoptic segmentation~\cite{kirillov2019panoptic}.
In this framework, a set of learnable object queries is refined through alternating self-attention among queries and cross-attention to image features in a Transformer decoder. Each processed query then predicts a class label via a linear layer and a binary segmentation mask via a dot product with the image features.
Several works have been built on top of this framework to advance the state of the art in universal image segmentation~\cite{cheng2022mask2former, yu2022kmaxdeeplab, jain2023oneformer, li2023maskdino, cavagnero2024pem}.
To obtain competitive results, they typically combine a pre-trained encoder, a pixel decoder for multi-scale feature fusion, and a Transformer decoder for query-based mask and class prediction.
When using large-scale pre-trained ViTs~\cite{oquab2023dinov2, simeoni2025dinov3} as encoders, models typically also include CNN-based adapter modules~\cite{chen2023vitadapter, xia2024vitcomer} to recover multi-scale features.

More recently, EoMT~\cite{kerssies2025eomt} challenged the reliance on such task-specific components by demonstrating that, with sufficiently large and well-trained ViTs, these components are largely redundant: by directly injecting learnable queries into the last encoder layers, EoMT achieves competitive accuracy at a significantly higher inference speed.
Moreover, EoMT benefits from ongoing ViT advancements, such as Token Merging~\cite{bolya2023tome,norouzi2024algm,lu2023cts} and FlashAttention~\cite{dao2024flashattention}.

\PAR{Video Segmentation.}
Video segmentation extends frame-level segmentation to the temporal domain, encompassing video instance segmentation (VIS)~\cite{yang2019vis}, video panoptic segmentation (VPS)~\cite{kim2020vps}, and video semantic segmentation (VSS)~\cite{nilsson2018vss}.
State-of-the-art methods~\cite{zhang2023dvis, zhang2025dvis++, zhou2024dvisdaq, lee2025cavis, GenVIS} follow a decoupled paradigm: a per-frame segmenter produces masks and object queries, while a separate tracker associates them across time using specialized components such as context-aware feature extractors, re-identification layers, and temporal Transformer layers.

VidEoMT~\cite{norouzi2026videomt} recently extended EoMT's encoder-only philosophy to video, showing that these tracking modules can be replaced by a lightweight query propagation mechanism used alongside the plain ViT encoder, achieving $5$--$10\times$ speedups while preserving competitive accuracy.
Crucially, both EoMT and VidEoMT require finetuning the full ViT encoder, preventing the encoder from being used for other downstream tasks.
Our work directly addresses this limitation by moving query processing outside the encoder: we introduce a lightweight decoder that applies EoMT's query-based attention mechanism on top of a fully frozen ViT, retaining the architectural simplicity and inference speed of the encoder-only approach while keeping the encoder features shareable across different downstream tasks.

\section{Method}
\label{sec:method}

\subsection{Preliminaries}
\label{sec:method:prelim}

\PARbegin{Vision Transformer.}
A Vision Transformer~\cite{dosovitskiy2021vit} partitions an image $\mathbf{I}\!\in\!\mathbb{R}^{3\times H\times W}$ into $N$ non-overlapping patches of size ${p\!\times\!p}$, which are linearly projected into patch tokens $\mathbf{X}^0\!\in\!\mathbb{R}^{D\times N}$.
These tokens are then refined by $L$ Transformer layers~\cite{vaswani2017attention}, where each layer~$l$ applies multi-head self-attention (\texttt{MHSA}) and a feed-forward network (\texttt{FFN}) with residual connections:
\begin{equation}
\begin{split}
    \mathbf{Z}^l &= \mathbf{X}^l + \texttt{MHSA}\!\left(\texttt{Norm}(\mathbf{X}^l)\right), \\
    \mathbf{X}^{l+1} &= \mathbf{Z}^l + \texttt{FFN}\!\left(\texttt{Norm}(\mathbf{Z}^l)\right).
\end{split}
\label{eq:vit}
\end{equation}
where \texttt{Norm} denotes Layer Normalization~\cite{ba2016layernorm}.
The final tokens $\mathbf{X}^L$ can be rearranged in a spatial grid to produce image features at $\frac{H}{p}\!\times\!\frac{W}{p}$ resolution.

\PAR{EoMT.} 
Traditional ViT-based segmentation models stack a CNN adapter~\cite{chen2023vitadapter}, a pixel decoder, and a Transformer decoder~\cite{cheng2022mask2former} with masked cross-attention on top of the ViT encoder.
EoMT~\cite{kerssies2025eomt} shows that, with strong VFM pre-training~\cite{oquab2023dinov2, simeoni2025dinov3} and sufficiently large model size, all these modules are largely redundant and can be removed, yielding large speedups at comparable accuracy.
Instead of leveraging these task-specific modules, $K$ learnable queries $\mathbf{Q}^\textrm{lrn}\!=\!\{\mathbf{q}^\textrm{lrn}_i\!\in\!\mathbb{R}^D\}_{i=1}^{K}$ are concatenated to the patch tokens after the first $L_1$ encoder layers. 
The last $L_2$ layers then process the patch tokens and queries jointly as a single sequence.
This means that the decoder's \texttt{MHSA} operations, in which all tokens attend to all others, simultaneously conducts query self-attention and query-to-patch attention without an explicit cross-attention mechanism as typically used in Transformer decoders~\cite{cheng2022mask2former}.
Finally, a lightweight mask module predicts, for each query, a class label using a linear layer and a segmentation mask through a dot product with patch tokens $\mathbf{X}^L$ spatially reorganized and upscaled to resolution $\frac{H}{4}\!\times\!\frac{W}{4}$.
During training, masked attention~\cite{cheng2022mask2former} is applied at each of the $L_2$ layers: an intermediate mask prediction is computed per query and used to restrict query-to-patch attention to the query's predicted mask region, improving training convergence.
However, in inference, predicting intermediate masks on every layer is expensive, and the custom attention pattern is incompatible with FlashAttention~\cite{dao2024flashattention}, which requires standard unmasked attention.
To resolve this, \textit{mask annealing}~\cite{kerssies2025eomt} gradually phases out masked attention during training, so the final model operates without any masking, unlocking FlashAttention, and roughly halving inference latency.
Crucially, for EoMT, the \emph{entire} encoder is finetuned, as the pre-trained attention must adapt to the newly introduced query tokens in order to produce meaningful query representations from which accurate segmentation predictions can be made.

\PAR{VidEoMT.}
VidEoMT~\cite{norouzi2026videomt} extends EoMT to online video segmentation by replacing all task-specific tracking components~\cite{zhang2023dvis, zhang2025dvis++, lee2025cavis} (tracker, context-aware features, re-identification layers) with a lightweight query-level temporal propagation mechanism.
In the first frame ($t\!=\!0$), the model operates identically to EoMT, taking learnable queries $\mathbf{Q}^\textrm{lrn}$ and producing output queries $\mathbf{Q}_0$.
In subsequent frames ($t\!>\!0$), the output queries from the previous frame are fused with the learned queries through element-wise addition after a linear projection:
\begin{equation}
    \mathbf{Q}_{t}^{\mathcal{F}} = \texttt{Linear}\!\left(\mathbf{Q}_{t-1}\right) + \mathbf{Q}^\textrm{lrn}.
\label{eq:query_fusion}
\end{equation}
The fused queries $\mathbf{Q}_{t}^{\mathcal{F}}$ are then fed into the encoder instead of the learnable queries, carrying temporal context from the previous frame while retaining the ability to detect newly appearing objects.
This design places all temporal reasoning in the query tokens and in the lightweight propagation module, avoiding slow task-specific modules.
However, as with EoMT, the full encoder is finetuned.

\begin{figure*}[t]
\vspace{-3pt}
\centering
\includegraphics[
    width=0.78\linewidth,
    keepaspectratio
]{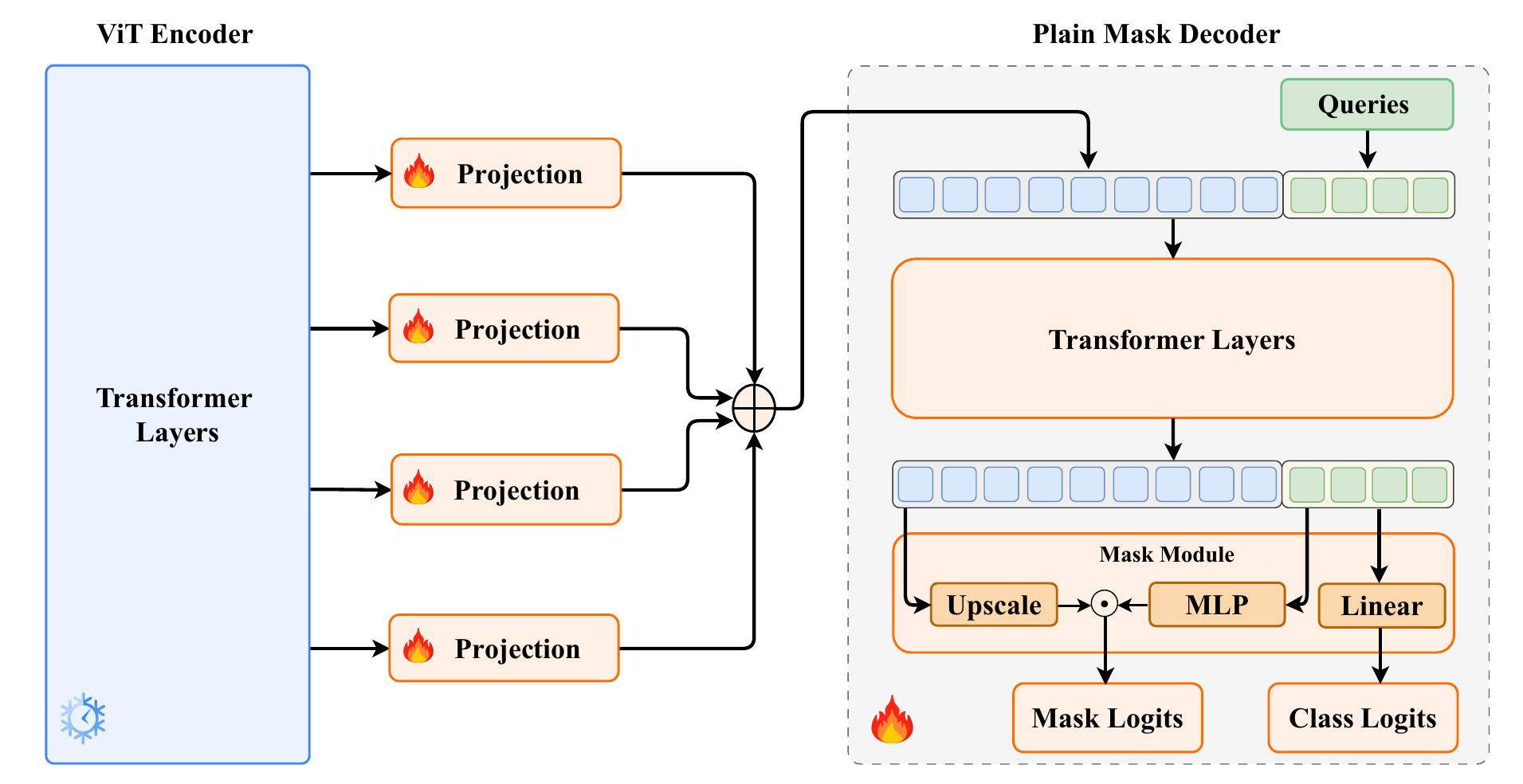}
\caption{\textbf{\ours{} (\oursShort{}) Architecture.} Instead of injecting the query tokens within the ViT encoder as in the encoder-only framework of EoMT and VidEoMT, we extract features at multiple encoder levels and feed them into an efficient segmentation decoder that processes queries and patch tokens in parallel. $\oplus$ denotes element-wise addition. $\odot$ denotes the dot product.}
\label{fig:arch}
\end{figure*}

\PAR{Incompatibility with Frozen Encoders.}
In both EoMT and VidEoMT, queries are concatenated with patch tokens \emph{inside} the encoder, after which both are processed by the \emph{same} attention weights.
Since pre-trained attention has no representation of these additional tokens, the weights must be updated for the joint patch-query attention to be effective.
As a result, freezing the encoder does not merely reduce accuracy, it fundamentally prevents the mechanism from functioning, as the frozen \texttt{MHSA} has no means to meaningfully attend to or from query tokens it was never trained on.
We empirically verify this in \cref{tab:steps_down}: with a frozen decoder, using an encoder-only model results in a complete collapse of performance, confirming the specific incompatibility of EoMT's joint patch-query attention with frozen weights.
The same behavior applies in the video domain.
Although this is not a limitation of the encoder-only design \textit{per se}, it does make encoder-only models incompatible with the frozen-encoder paradigm for which large-scale VFMs are currently being designed~\cite{simeoni2025dinov3}.

\subsection{\ours{} Architecture}
\label{sec:method:decoder}

To enable efficient and accurate segmentation while using a frozen ViT encoder, our key insight is that the behavior of the last $L_2$ layers of EoMT's encoder can be mimicked by small decoder consisting of the same types of layers.
The application of this decoder, which we call the \decoder{} (\decoderShort{}), allows the ViT encoder to remain frozen, preserving pre-training knowledge and enabling the ViT to serve any number of downstream tasks in parallel. The resulting model is called the \ours{} (\oursShort{}).

\PARbegin{Decoder Layers.}
The decoder contains vanilla Transformer layers that mirror the architecture of the DINOv3 encoder layers~\cite{simeoni2025dinov3}, matching the hidden dimension, the number of heads and the \texttt{FFN} design.
Following EoMT, queries $\mathbf{Q}^\textrm{lrn}$ and patch tokens $\mathbf{X}^L$ are concatenated into a single sequence and processed through standard \texttt{MHSA}: since all tokens attend to all others, this jointly achieves query self-attention, patch token self-attention, and attention between queries and patches in both directions, just like EoMT.
Unlike EoMT, however, these are standalone layers trained from randomly initialized weights, while the encoder weights remain entirely frozen.
We use $L_d\!=\!6$ decoder layers by default (see \cref{tab:decoder_layers}).

Following EoMT, masked attention~\cite{cheng2022mask2former} is applied during training in each decoder layer, forcing each query to attend to its predicted region.
Mask annealing~\cite{kerssies2025eomt} then progressively phases out this masking, yielding a consistently faster mask-free decoder at inference.
We denote the set of decoder output queries as $\hat{\mathbf{Q}} = \{\hat{\mathbf{q}}_i \in \mathbb{R}^D\}_{i=1}^K$ and the decoder output patch tokens as $\hat{\mathbf{X}} \in \mathbb{R}^{D \times N}$.

\PAR{Mask Module.}
We adopt the same prediction head as EoMT and VidEoMT.
For each output query $\hat{\mathbf{q}}_i \in \hat{\mathbf{Q}}$, class logits $\mathbf{c}_i \in \mathbb{R}^C$ are predicted by a linear layer.
Mask logits $\mathbf{M}_i \in \mathbb{R}^{\frac{H}{4}\times\frac{W}{4}}$ are obtained by first applying a three-layer \texttt{MLP} to $\hat{\mathbf{q}}_i$, then taking its dot product with ${\mathbf{X}^\textrm{up}} \in \mathbb{R}^{D \times \frac{H}{4} \times \frac{W}{4}}$, obtained after reshaping and upscaling $\hat{\mathbf{X}}$.

\PAR{Lateral Connections.} When EoMT is finetuned, the early $L_1$ encoder layers can adapt to the target task producing features that are more useful for the query-processing layers that follow.
A frozen encoder cannot do this: its features are fixed and not adapted to the target task. As a consequence, the final encoder output $\mathbf{X}^L$ may not contain all cues that are useful for segmentation (\eg, edges, boundaries), even though these may be available in features of earlier layers.

Analogously to how adapters~\cite{chen2023vitadapter} extract multi-scale features at selected encoder depths, we introduce lateral connections that collect patch tokens $\mathbf{X}^l$ from evenly spaced encoder layers (including the final layer~$L$).
This allows the decoder to leverage the rich information available at different encoder depths, improving the quality of the mask predictions.
For normalization, we follow the strategy that the DINOv3~\cite{simeoni2025dinov3} paper uses when feeding intermediate features to the decoder for downstream tasks. Specifically, we apply the encoder's final Layer Normalization~\cite{ba2016layernorm} to all extracted token features, followed by a trainable Batch Normalization~\cite{ioffe2015batchnorm} layer.
All token features are then projected by a two-layer \texttt{MLP} with a residual connection. Features from all branches are summed element-wise into a single multi-depth representation that serves as the set of patch tokens that is fed into the decoder.

\PAR{Positional Encoding.} DINOv3~\cite{simeoni2025dinov3} employs Rotary Position Embeddings (RoPE)~\cite{su2024roformer} in every encoder layer, encoding relative spatial positions directly into the attention computation by rotating queries and keys before the dot product.
Since our decoder uses its own freshly initialized attention layers, applying the same RoPE provides them with explicit spatial context, supplementing the positional information already embedded in the patch tokens.

We therefore apply RoPE to the decoder layers: patch tokens retain the grid coordinates assigned by the encoder, while query tokens receive no positional encoding, as keeping them position-free preserves the permutation-invariant nature of the set of object queries. Their spatial grounding is provided implicitly through attention to positioned patches and through the mask module.
This adds no learnable parameters, as RoPE is a deterministic function of position.

\subsection{Temporal Modeling}
\label{sec:method:video}

Our approach extends naturally to online video segmentation by adopting VidEoMT's query propagation mechanism~\cite{norouzi2026videomt}.
Each frame is processed independently by the frozen encoder and lateral connections; the only temporal link is found in the queries fed to the decoder.

At $t\!=\!0$ the decoder receives the learnable queries $\mathbf{Q}^\textrm{lrn}$.
In subsequent frames ($t\!>\!0$), the output queries of the previous frame are fused with the learnable queries following \cref{eq:query_fusion} and fed to the \decoderShort{} decoder.
As in VidEoMT, no tracker, re-identification layers, or context-aware features are needed, yielding a simple and fast video architecture.

\section{Experiments}
\label{sec:exp}

\subsection{Experimental Setup}

Unless stated otherwise, we follow the experimental setup of EoMT~\cite{kerssies2025eomt} for image-level experiments and VidEoMT~\cite{norouzi2026videomt} for video-level experiments.

\PARbegin{Datasets.}
For image segmentation, we use COCO~\cite{lin2014coco} for panoptic and instance segmentation, and ADE20K~\cite{zhou2017ade20k} for semantic segmentation.
For video, we use YouTube-VIS 2019 and 2021~\cite{yang2019video} for video instance segmentation (VIS), VIPSeg~\cite{miao2022large} for video panoptic segmentation (VPS), and VSPW~\cite{miao2021vspw} for video semantic segmentation (VSS).

\PAR{Models.}
Unless stated otherwise, we use DINOv3-L~\cite{simeoni2025dinov3} as the encoder with a patch size of $16\!\times\!16$ and a $640\!\times\!640$ resolution.
Our \decoderShort{} matches the hidden dimension of the encoder to avoid information bottlenecks. Register and class tokens are propagated together with the patch tokens through the decoder layers, and the \texttt{MLP} expansion factors are set to 1.
For ViT-Adapter, we follow the implementation from the DINOv3 paper~\cite{simeoni2025dinov3}, removing the injectors to preserve the frozen encoder representation and adding a learnable Batch Normalization~\cite{ioffe2015batchnorm} layer at each feature level.
At inference, the \texttt{QKV} projections within the \texttt{MHSA} layers of the encoder and decoder are fused to reduce latency.

\PAR{Training.}
All models are trained in mixed precision with AdamW~\cite{loshchilov2019adamw} and mask annealing following EoMT with DINOv3.
For image models, we use a batch size of 16, a learning rate of $2 \times 10^{-4}$, cosine annealing scheduler~\cite{loshchilov2017sgdr}, and training schedules of 12 and 16 epochs on COCO and ADE20K, respectively.
For video models, we use a batch size of 8 with a 5-frame temporal window, a learning rate of $10^{-4}$, polynomial learning rate decay with a power of 0.9, and training iterations and resolutions following CAVIS~\cite{lee2025cavis}. In both settings, we apply linear warmup for 6000 iterations.
Following common practice~\cite{kerssies2025eomt, norouzi2026videomt, cheng2022mask2former}, we use cross-entropy for classification and binary cross-entropy combined with Dice loss~\cite{milletari2016dice} for mask prediction.
For video, ground-truth matching follows VidEoMT: each object is matched to a query in its first appearing frame, and the assignment persists across subsequent frames. 

\PAR{Evaluation.}
For image segmentation, we report Panoptic Quality (PQ)~\cite{kirillov2019panoptic} for panoptic, mean Intersection over Union (mIoU) for semantic, and Average Precision (AP)~\cite{lin2014coco} for instance segmentation.
For video segmentation, we report AP and Average Recall (AR)~\cite{yang2019video} for VIS, Video Panoptic Quality (VPQ)~\cite{kim2020vps} and Segmentation and Tracking Quality (STQ)~\cite{weber2021step} for VPS, and mIoU along with Video Consistency (mVC)~\cite{miao2021vspw} for VSS.
Inference speed (FPS) is measured on an NVIDIA H100 GPU with FlashAttention-2~\cite{dao2024flashattention} and \texttt{torch.compile}~\cite{ansel2024pytorch2} enabled, using a batch size of 1 and automatic mixed precision. 
We set \texttt{torch.compile} to \textit{max-autotune} mode for image experiments and use the default settings for video experiments due to varying resolutions. GFLOPs are computed using \textit{fvcore}~\cite{metaresearch2023fvcore}. Both FPS and GFLOPs are measured by averaging across the validation sets. 

\subsection{Main Results}
\label{sec:res:main}

\begin{table}[t]
    \centering
    \footnotesize
    \setlength{\tabcolsep}{2.3pt}

    \begin{tabularx}{\linewidth}{
        clcc c c
        }
        \toprule
        
        & 
        Method & Params & GFLOPs & FPS & PQ \\
        
        \midrule

        \tablestep{0} & 
        ViT-Adapter + Mask2Former &
        340M &
        804 & 
        48 &
        56.4 \\

        \midrule

        \tablestep{1} &
        \downrightarrow w/o ViT-Adapter & 
        341M & 
        740 & 
        62 &
        54.0 \\

        \tablestep{2} &
        \downrightarrow w/o Pixel decoder & 
        336M & 
        686 & 
        77 &
        52.9 \\
        
        \tablestep{3} & 
        \downrightarrow  w/o Multi-scale & 
        327M & 
        673 & 
        81 &
        52.0 \\

        \tablestep{4} & 
        \downrightarrow w/o Transformer decoder = EoMT &
        315M & 
        668 & 
        162 &
        6.8 \\

        \bottomrule
    
    \end{tabularx}

\caption{\textbf{From ViT-Adapter + Mask2Former to EoMT.} Stepwise removal of Mask2Former modules toward EoMT with \textit{frozen} DINOv3~\cite{simeoni2025dinov3} encoder. Evaluated on COCO \textit{val2017}~\cite{lin2014coco}.}
\label{tab:steps_down}

\end{table}

\PARbegin{From ViT-Adapter + Mask2Former to EoMT.}
In \cref{tab:steps_down}, we evaluate the stepwise removal of the task-specific components from the state-of-the-art ViT-Adapter + Mask2Former~\cite{chen2023vitadapter, cheng2022mask2former} model toward the EoMT encoder-only design with a frozen DINOv3 encoder.
Steps \tablestep{1}--\tablestep{3} strip the adapter, pixel decoder, and multi-scale processing successively, dropping PQ from 56.4 to 52.0 while nearly doubling FPS. Unlike in the original EoMT results where these modules proved largely redundant under full encoder fine-tuning, the frozen setting exposes a clear accuracy loss: the remaining trainable parameters lack the capacity to compensate for the removed components when the encoder cannot adapt.
Step \tablestep{4} reaches the encoder-only EoMT configuration and, despite doubling again the speed, causes a far more severe collapse to 6.8~PQ.
Interestingly, this drop is disproportionate compared to any earlier step, highlighting the failure of joint patch-query attention when keeping weights frozen, as discussed in \cref{sec:method}.

\begin{table}
    \centering
    \footnotesize
    \setlength{\tabcolsep}{4pt}

    \begin{tabularx}{\linewidth}{cl Y YYYY}
        \toprule

        &
        {Method} & {Params} & {GFLOPs} & {FPS} & {PQ} \\

        \midrule
        \multicolumn{6}{c}{\textit{DINOv3 pre-training}} \\
        \midrule

        \tablestep{0} &
        EoMT~\cite{kerssies2025eomt} &
        316M & 668 & 162 & 6.8 \\

        \tablestep{1} &
        \downrightarrow\; w/ plain decoder &
        354M & 762 & 145 & 53.7 \\

        \tablestep{2} &
        \downrightarrow\; w/ lateral connections &
        357M & 767 & 142 & 55.9 \\

        \tablestep{3} &
        \downrightarrow\; w/ RoPE = \textbf{\oursShort} (Ours) &
        357M & 767 & 141 & 56.1 \\

        \midrule
        \multicolumn{6}{c}{\textit{DINOv2 pre-training}} \\
        \midrule

        \tablestep{0} &
        EoMT~\cite{kerssies2025eomt} &
        317M & 668 & 172 & 22.3 \\

        \tablestep{1} &
        \downrightarrow\; w/ plain decoder &
        356M & 762 & 150 & 51.4 \\

        \tablestep{2} &
        \downrightarrow\; w/ lateral connections &
        359M & 767 & 148 & 55.0 \\

        \tablestep{3} &
        \downrightarrow\; w/ RoPE = \textbf{\oursShort} (Ours) &
        359M & 767 & 146 & 55.5 \\

        \bottomrule

    \end{tabularx}

\caption{\textbf{From EoMT to \oursShort.} Stepwise addition of modules from EoMT with \textit{frozen} DINOv3~\cite{simeoni2025dinov3} and DINOv2~\cite{oquab2023dinov2} encoders to our \oursShort. Evaluated on COCO \textit{val2017}~\cite{lin2014coco}.}
\label{tab:steps_up_stacked}

\end{table}

\PAR{From EoMT with Frozen Encoder to \oursShort.}
In \cref{tab:steps_up_stacked}, we build up from the collapsed frozen EoMT baseline to our \oursShort{}.
Step \tablestep{1} adds a standalone \decoderShort{} and immediately closes most of the gap, reaching 53.7~PQ with DINOv3 while only reducing inference speed by a small margin.
Step \tablestep{2} incorporates lateral connections that fuse patch tokens collected at evenly spaced encoder depths, adding a further 2.2~PQ to reach 55.9~PQ at very little computational cost.
Step \tablestep{3} applies RoPE to the decoder, yielding the full \oursShort{} model at 56.1~PQ, a $+$0.2~PQ gain at virtually no cost, as RoPE introduces minimal latency and no additional learnable parameters.
Comparable trends hold for DINOv2, although the gains from lateral connections and RoPE over the plain decoder are more pronounced, with $+$3.6 and $+$0.5~PQ, respectively. This is consistent with EoMT's observation~\cite{kerssies2025eomt} that stronger pre-training reduces the necessity for additional components.
Overall, these results show that processing queries with our lightweight decoder and lateral connections is sufficient to recover almost all of the performance of the baseline with complex task-specific modules, while being $3\times$ faster (\emph{cf.}~\cref{tab:steps_down}).

\begin{table}[t]
\centering
\footnotesize
\setlength{\tabcolsep}{1.0pt}
\begin{tabularx}{\linewidth}{
lcYYYYYY
}
\toprule

Model && 
Pre-train & 
Params &
GFLOPs & 
FPS & 
PQ \\

\midrule

ViT-Adapter + M2F &&
\multirow{2}{*}{DINOv3} & 
340M &
804 &
48 &
56.4 \\

\textbf{\oursShort} (Ours)  &&
 &
357M &
767 &
141 &
56.1
\\

\midrule

ViT-Adapter + M2F && 
\multirow{2}{*}{DINOv2} &
342M &
804 &
49 &
55.7  
\\

\textbf{\oursShort} (Ours)  &&
 &
359M &
767 &
146 &
55.5 \\

\midrule

ViT-Adapter + M2F && 
\multirow{2}{*}{IN21K} &
342M &
804 &
49 &
51.0 \\

\textbf{\oursShort} (Ours)  &&
 &
359M &
767 &
146 &
49.7 \\

\midrule

ViT-Adapter + M2F &&
\multirow{2}{*}{IN1K} & 
342M &
804 &
49 &
47.0 \\

\textbf{\oursShort} (Ours)  &&
 &
359M &
767 &
146 &
44.6 \\

\bottomrule

\end{tabularx}
\caption{\textbf{Pre-training.} Impact of pre-training on segmentation quality. Evaluated on COCO \textit{val2017}~\cite{lin2014coco}.
}
\label{tab:pretraining}
\end{table}

\PAR{Impact of Pre-training.}
Next, we examine how the extent of pre-training affects the performance of \oursShort{} in \cref{tab:pretraining}.
We evaluate DINOv3~\cite{simeoni2025dinov3} and DINOv2~\cite{oquab2023dinov2} as large-scale self-supervised models, as well as encoders trained in supervised fashion on ImageNet-21K~\cite{deng2009imagenet} and ImageNet-1K~\cite{russakovsky2015imagenet}.
The results show that large-scale pre-training is essential for \oursShort{} to close the gap with ViT-Adapter + M2F.
The accuracy difference with DINOv3 and DINOv2 is just 0.3 and 0.2 PQ, respectively, while the gap increases to 1.3 PQ under ImageNet-21K supervision and 2.4 PQ under ImageNet-1K.
The widening gap at less extensive pre-training confirms that the frozen encoder must carry sufficiently rich features for the lightweight \decoderShort{} to substitute complex task-specific modules, which is in line with the scaling observations of EoMT~\cite{kerssies2025eomt} and VidEoMT~\cite{norouzi2026videomt}.

\begin{table}[t]
\centering
\footnotesize
\setlength{\tabcolsep}{4pt}
\begin{tabularx}{\linewidth}{lYYYYY}
\toprule

Model &
Size &
Params &
GFLOPs &
FPS &
PQ \\

\midrule

ViT-Adapter + M2F &
\multirow{2}{*}{L} &
340M & 804 & 48 & 56.4 \\

\textbf{\oursShort} (Ours) &
& 357M & 767 & 141 & 56.1 \\

\midrule

ViT-Adapter + M2F &
\multirow{2}{*}{B} &
116M & 334 & 64 & 53.6 \\

\textbf{\oursShort} (Ours) &
& 116M & 281 & 262 & 52.8 \\

\midrule

ViT-Adapter + M2F &
\multirow{2}{*}{S} &
45M & 163 & 73 & 49.6 \\

\textbf{\oursShort} (Ours) &
& 29M & 91 & 400 & 45.3 \\

\bottomrule

\end{tabularx}
\caption{\textbf{Model Size.} Impact of model size on segmentation quality and inference speed. Evaluated on COCO \textit{val2017}~\cite{lin2014coco}.}
\label{tab:model_size}
\end{table}

\begin{table}[t]
\centering
\footnotesize
\setlength{\tabcolsep}{2.0pt}
\begin{tabularx}{0.8\linewidth}{
YYYYYc
}
\toprule

Blocks &
Params &
GFLOPs & 
FPS & 
PQ \\

\midrule

2 &
332M &
695 &
153 &
54.0 \\

4 &
344M &
731 &
146 &
55.6 \\

6 &
357M &
767 &
141 &
56.1 \\

8 &
370M &
803 &
133 &
56.1 \\

\bottomrule

\end{tabularx}
\caption{\textbf{Decoder Layers.} Impact of number of decoder layers on segmentation quality and inference speed. Evaluated on COCO \textit{val2017}~\cite{lin2014coco}.
}
\label{tab:decoder_layers}
\end{table}

\PAR{Impact of Model Size.}
\cref{tab:model_size} extends the comparison to different model sizes.
While the accuracy gap between \oursShort{} and ViT-Adapter + M2F reaches 4.3~PQ with the small ViT-S model, it narrows to 0.8~PQ with ViT-B and decreases to just 0.3~PQ with ViT-L.
This scaling pattern confirms that \decoderShort{} requires a frozen encoder with sufficient capacity to compensate for the absence of dedicated feature-adaptation modules.
Importantly, these results also show that \oursShort{} delivers considerably higher inference speed across all sizes, with ${\sim}3\times$ at ViT-L. 
In fact, ViT-L is even faster than ViT-Adapter + M2F at ViT-S at much greater accuracy, demonstrating \oursShort{}'s superior accuracy--efficiency trade-off.

\begin{table*}[t]
    \centering
    \footnotesize
    \renewcommand{\tabcolsep}{9pt}
    \begin{tabularx}{\linewidth}
    {ll cccccc cc}
        \toprule
        
        Method &
        Backbone &
        Pre-training &
        Encoder &
        Input size &
        Params &
        GFLOPs & FPS & PQ & AP \\

        \midrule

        Mask2Former$^\dagger$~\cite{cheng2022mask2former} & 
        Swin-L~\cite{liu2021swin} &
        IN21K & \twemoji{fire} &
        $800^2$ &
        216M &
        868 & 24 & 57.8 & 50.1 \\
        
        kMaX-DeepLab~\cite{yu2022kmaxdeeplab} & 
        ConvNext-L~\cite{liu2022convnext} & 
        IN21K & \twemoji{fire} &
        $1281^2$ &
        232M &
        -- & -- & 58.0 & -- \\
        
        OneFormer$^\dagger$~\cite{jain2023oneformer} & 
        DiNAT-L~\cite{hassani2022dinat} & 
        IN21K & \twemoji{fire} &
        $800^2$ &
        223M &
        736 & 20 & 58.0 & 49.2 \\
        
        MaskDINO$^\dagger$~\cite{li2023maskdino} & 
        Swin-L~\cite{liu2021swin} & 
        IN21K & \twemoji{fire} &
        $800^2$ &
        223M &
        1326 & 14 & 58.3 & 52.3 \\

        EoMT~\cite{kerssies2025eomt} & 
        ViT-L~\cite{dosovitskiy2021vit} & 
        DINOv2 & \twemoji{fire} &
        $640^2$ &
        316M &
        669 & 172 & 56.0 & 45.2 \\

        EoMT~\cite{kerssies2025eomt} & 
        ViT-L~\cite{dosovitskiy2021vit} & 
        DINOv3 & \twemoji{fire} &
        $640^2$ &
        315M &
        668 & 162 & 56.8 & 45.9 \\

        EoMT~\cite{kerssies2025eomt} & 
        ViT-L~\cite{dosovitskiy2021vit} & 
        DINOv2 & \twemoji{fire} &
        $1280^2$ &
        322M &
        4146 & 35 & 58.3 & 48.8 \\

        EoMT~\cite{kerssies2025eomt} & 
        ViT-L~\cite{dosovitskiy2021vit} & 
        DINOv3 & \twemoji{fire} &
        $1280^2$ &
        315M &
        4144 & 34 & 58.9 & 49.9 \\

        \midrule

        Mask2Former~\cite{cheng2022mask2former} & 
        ViT-Adapter-L~\cite{chen2023vitadapter} & 
        DINOv2 & \twemoji{2744} &
        $640^2$ &
        342M &
        804 & 49 & 55.7 & 45.0 \\

        Mask2Former~\cite{cheng2022mask2former} & 
        ViT-Adapter-L~\cite{chen2023vitadapter} & 
        DINOv3 & \twemoji{2744} &
        $640^2$ &
        340M &
        804 & 48 & 56.4 & 45.9 \\

        \textbf{\oursShort} (Ours) & 
        ViT-L~\cite{dosovitskiy2021vit} & 
        DINOv2 & \twemoji{2744} &
        $640^2$ &
        359M &
        767 & 146 & 55.5 & 44.4 \\

        \textbf{\oursShort} (Ours) & 
        ViT-L~\cite{dosovitskiy2021vit} & 
        DINOv3 & \twemoji{2744} &
        $640^2$ &
        357M &
        767 & 141 & 56.1 & 45.4 \\

        \midrule

        Mask2Former~\cite{cheng2022mask2former} & 
        ViT-Adapter-L~\cite{chen2023vitadapter} & 
        DINOv2 & \twemoji{2744} &
        $1280^2$ &
        347M &
        4711 & 16 & 56.6 & 47.6 \\

        Mask2Former~\cite{cheng2022mask2former} & 
        ViT-Adapter-L~\cite{chen2023vitadapter} & 
        DINOv3 & \twemoji{2744} &
        $1280^2$ &
        340M &
        4711 & 15 & 58.1 & 50.0 \\

        \textbf{\oursShort} (Ours) & 
        ViT-L~\cite{dosovitskiy2021vit} & 
        DINOv2 & \twemoji{2744} &
        $1280^2$ &
        364M &
        4925 & 30 & 56.1 & 45.9 \\
        
        \textbf{\oursShort} (Ours) & 
        ViT-L~\cite{dosovitskiy2021vit} & 
        DINOv3 & \twemoji{2744} &
        $1280^2$ &
        357M &
        4925 & 29 & 58.1 & 49.0 \\

        \bottomrule
    \end{tabularx}
    \caption{\textbf{\oursShort{} for Panoptic and Instance Segmentation on COCO \textit{val2017}~\cite{lin2014coco}.} $^\dagger$During inference, these models resize the shortest side of images to the indicated scale, while preserving the aspect ratio.
    }
    \label{tab:sota_coco}
\end{table*}

\begin{table*}[t]
    \centering
    \footnotesize
    \renewcommand{\tabcolsep}{11pt}
    \begin{tabularx}{\linewidth}
    {llccc cccc}
        \toprule
        
        Method &
        Backbone &
        Pre-training &
        Encoder &
        Input size &
        Params &
        GFLOPs & FPS & mIoU \\

        \midrule

        Mask2Former$^\dagger$~\cite{cheng2022mask2former} & 
        Swin-L~\cite{liu2021swin} &
        IN21K & \twemoji{fire} &
        $640^2$ &
        216M &
        -- & 33 & 56.1 \\

        MaskDINO$^\dagger$~\cite{li2023maskdino} & 
        Swin-L~\cite{liu2021swin} & 
        IN21K & \twemoji{fire} &
        $640^2$ &
        223M &
        -- & -- & 56.6 \\
        
        OneFormer$^\dagger$~\cite{jain2023oneformer} & 
        ConvNext-XL~\cite{liu2022convnext} & 
        IN21K & \twemoji{fire} &
        $640^2$ &
        373M &
        607 & 21 & 57.4 \\
        
        OneFormer$^\dagger$~\cite{jain2023oneformer} & 
        DiNAT-L~\cite{hassani2022dinat} & 
        IN21K & \twemoji{fire} &
        $896^2$ &
        223M &
        678 & 19 & 58.1 \\

        EoMT~\cite{kerssies2025eomt} & 
        ViT-L~\cite{dosovitskiy2021vit} & 
        DINOv2 & \twemoji{fire} &
        $512^2$ &
        316M &
        721 & 154 & 58.4 \\

        EoMT~\cite{kerssies2025eomt} & 
        ViT-L~\cite{dosovitskiy2021vit} & 
        DINOv3 & \twemoji{fire} &
        $512^2$ &
        315M &
        721 & 151 & 59.5 \\
        
        \midrule
        
        Mask2Former~\cite{cheng2022mask2former} & 
        ViT-Adapter-L~\cite{chen2023vitadapter} & 
        DINOv2 & \twemoji{2744} &
        $512^2$ &
        341M &
        879 & 41 & 57.5 \\

        \textbf{\oursShort} (Ours)  & 
        ViT-L~\cite{dosovitskiy2021vit} & 
        DINOv2 & \twemoji{2744} &
        $512^2$ &
        358M &
        823 & 135 & 57.2 \\

        \midrule

        Mask2Former~\cite{cheng2022mask2former} & 
        ViT-Adapter-L~\cite{chen2023vitadapter} & 
        DINOv3 & \twemoji{2744} &
        $512^2$ &
        340M &
        879 & 40 & 58.7 \\

        \textbf{\oursShort} (Ours)  & 
        ViT-L~\cite{dosovitskiy2021vit} & 
        DINOv3 & \twemoji{2744} &
        $512^2$ &
        357M &
        823 & 128 & 58.5 \\
        
        \bottomrule
    \end{tabularx}
    \caption{\textbf{\oursShort{} for Semantic Segmentation on ADE20K~\cite{zhou2017ade20k}.} $^\dagger$These models resize the shortest side of images to the indicated scale during inference, while preserving the aspect ratio.}
    \label{tab:sota_semantic}
\end{table*}

\PAR{Number of Decoder Layers.}
In \cref{tab:decoder_layers}, we evaluate how the depth of the decoder affects accuracy and speed.
Increasing from 2 to 4 layers brings a consistent improvement from 54.0 to 55.6~PQ, and 6 layers add a further 0.5~PQ to reach 56.1~PQ.
Beyond that, scaling up to 8 layers yields no additional gain.
Performance thus saturates quickly with depth, showing that a shallow decoder is sufficient for this design. We therefore set $L_d\!=\!6$ as the default.

\subsection{Image Segmentation Benchmarks}

\PARbegin{Panoptic and Instance Segmentation.}
In \cref{tab:sota_coco}, we compare \oursShort{} with state-of-the-art panoptic and instance segmentation methods on COCO.
Compared to the state-of-the-art frozen-encoder method that uses task-specific components (ViT-Adapter + Mask2Former), \oursShort{} obtains comparable performance at much higher speed.
\oursShort{} scores only 0.3~PQ and 0.5~AP lower at $640^2$ resolution, but is ${\sim}3\times$ faster. 
This demonstrates \oursShort{}'s effectiveness and efficiency.
Compared to the encoder-only EoMT, which \textit{does} fully finetune the encoder, \oursShort{} obtains only slightly lower accuracy and speeds, with a 0.7~PQ difference and 13\% speed decrease at $640^2$. 
Remarkably, \oursShort{} achieves this while keeping the encoder frozen, making it much more suitable for applications that require reusing the frozen encoder for multiple downstream tasks.

\PAR{Semantic Segmentation.}
In \cref{tab:sota_semantic}, we evaluate \oursShort{} for semantic segmentation on ADE20K.
Compared to ViT-Adapter + Mask2Former with a frozen encoder, \oursShort{} obtains almost the same segmentation accuracy, with drops of only 0.3 and 0.2 mIoU in combination with DINOv2 and DINOv3, respectively.
Once again, \oursShort{} is considerably faster, with inference speed gains of more than $3\times$, at 128 \vs 40 FPS.
Like for panoptic and instance segmentation, EoMT with a finetuned encoder performs slightly better than \oursShort{}, but the accuracy drop remains modest at 1.0 or 1.2 mIoU.
Of course, \oursShort{} obtains these scores without finetuning the encoder, improving its applicability.
Together, these results confirm that the earlier observed accuracy and efficiency advantages also extend to semantic segmentation, demonstrating the generality of \oursShort{}'s design.

 \begin{table*}[t!]
    \centering
    \scriptsize
    \renewcommand{\tabcolsep}{2.5pt}
    \begin{tabularx}{\linewidth}
    {llcc c YYYYY c YYYYY}
    \toprule
    \multirow{2}[2]{*}{Method} &
    \multirow{2}[2]{*}{Backbone} &
    \multirow{2}[2]{*}{Pre-training} &
    \multirow{2}[2]{*}{Encoder} && 
    \multicolumn{5}{c}{YouTube-VIS 2019 \textit{val}~\cite{yang2019video}} &&
    \multicolumn{5}{c}{YouTube-VIS 2021 \textit{val}~\cite{yang2019video}} \\
    \cmidrule{6-10} \cmidrule{12-16}
    &&&&&
    GFLOPs & FPS & AP & AP\textsubscript{75} & AR\textsubscript{10} &&
    GFLOPs & FPS & AP & AP\textsubscript{75} & AR\textsubscript{10} \\
    \midrule

    MinVIS~\cite{huang2022minvis} & Swin-L~\cite{liu2021swin} & IN21K & \twemoji{fire} &&
    401 & 29 & 61.6 & 68.6 & 66.6 && %
    255 & 30 & 55.3 & 62.0 & 60.8 %
    \\
    DVIS~\cite{zhang2023dvis} & Swin-L~\cite{liu2021swin} & IN21K & \twemoji{fire} &&
    411 & 23 & 63.9 & 70.4 & 69.0 && %
    405 & 24 & 58.7 & 66.6 & 64.6 %
    \\
    DVIS-DAQ~\cite{zhou2024dvisdaq} & Swin-L~\cite{liu2021swin} & IN21K & \twemoji{fire} &&
    415 & 13 & 65.7 & 73.6 & 70.7 && %
    410 & 11 & 61.1 & 68.2 & 66.6 %
    \\
    DVIS++~\cite{zhang2025dvis++} & ViT-Adapter-L~\cite{chen2023vitadapter} & DINOv2 & \twemoji{fire} &&
    846 & 18 & 67.7 & 75.3 & 73.7 && %
    830 & 17 & 62.3 & 70.2 & 68.0 %
    \\
    DVIS-DAQ~\cite{zhou2024dvisdaq} & ViT-Adapter-L~\cite{chen2023vitadapter} & DINOv2 & \twemoji{fire} &&
    851 & 10 & 68.3 & 76.1 & 73.5 && %
    836 & 10 & 62.4 & 70.8 & 68.0 %
    \\
    CAVIS~\cite{lee2025cavis} & ViT-Adapter-L~\cite{chen2023vitadapter} & DINOv2 & \twemoji{fire} &&
    838 & 15 & 68.9 & 76.2 & 73.6 && %
    824 & 15 & 64.6 & 72.5 & 69.3 %
    \\
    VidEoMT~\cite{norouzi2026videomt} & ViT-L~\cite{dosovitskiy2021vit} & DINOv2 & \twemoji{fire} &&
    566 & 160 & 68.6 & 75.6 & 73.9 && %
    560 & 160 & 63.1 & 69.3 & 68.1 %
    \\
    VidEoMT~\cite{norouzi2026videomt} & ViT-L~\cite{dosovitskiy2021vit} & DINOv3 & \twemoji{fire} &&
    566 & 133 & 68.9 & 77.4 & 74.8 && %
    560 & 133 & 63.2 & 71.6 & 69.1 %
    \\
    
    \midrule
    CAVIS~\cite{lee2025cavis} & ViT-Adapter-L~\cite{chen2023vitadapter} & DINOv2 & \twemoji{2744} &&
    838 & 15 & 68.5 & 75.8 & 73.5 && %
    824 & 15 & 64.3 & 72.0 & 68.9 %
    \\
    \textbf{\oursShort} (Ours)  & ViT-L~\cite{dosovitskiy2021vit} & DINOv2 & \twemoji{2744} &&
    617 & 124 & 68.8 & 75.2 & 73.9 && %
    616 & 124 & 63.8 & 69.4 & 68.1 %
    \\

    \midrule
   CAVIS~\cite{lee2025cavis} & ViT-Adapter-L~\cite{chen2023vitadapter} & DINOv3 & \twemoji{2744} &&
    838 & 13 & 68.8 & 75.6 & 73.3 && %
    824 & 13 & 63.9 & 71.6 & 68.2 %
    \\
    \textbf{\oursShort} (Ours)  & ViT-L~\cite{dosovitskiy2021vit} & DINOv3 & \twemoji{2744} &&
    617 & 113 & 69.2 & 76.5 & 74.6 && %
    616 & 113 & 64.3 & 71.2 & 69.0 %
    \\

    \bottomrule
    \end{tabularx}
    \caption{\textbf{\oursShort{} for Online VIS on YouTube-VIS 2019 and 2021~\cite{yang2019video}.}}
    \label{tab:sota_comparison_ytvis19_ytvis21}
\end{table*}

\begin{table}[t]
    \centering
    \scriptsize
    \renewcommand{\tabcolsep}{2pt}
    \begin{tabularx}{\linewidth}{llcc c cYYY}
    \toprule
    \multirow{2}[2]{*}{Method} &
    \multirow{2}[2]{*}{Backbone} &
    \multirow{2}[2]{*}{PT} &
    \multirow{2}[2]{*}{Encoder} && 
    \multicolumn{4}{c}{VIPSeg \textit{val}~\cite{miao2022large}} \\
    \cmidrule{6-9}
    &&&&&
    GFLOPs & FPS & VPQ & STQ \\
    \midrule

    DVIS~\cite{zhang2023dvis} & Swin-L~\cite{liu2021swin} & I & \twemoji{fire} &
    & 879 & 20 & 54.7 & 47.7 \\

    DVIS++~\cite{zhang2025dvis++} & ViT-Adapter-L~\cite{chen2023vitadapter} & D2 & \twemoji{fire} &
    & 2290 & 13 & 56.0 & 49.8 \\

    DVIS-DAQ~\cite{zhou2024dvisdaq} & ViT-Adapter-L~\cite{chen2023vitadapter} & D2 & \twemoji{fire} &
    & 2315 & 4 & 57.4 & 52.0 \\

    CAVIS~\cite{lee2025cavis} & ViT-Adapter-L~\cite{chen2023vitadapter} & D2 & \twemoji{fire} &
    & 2612 & 10 & 56.9 & 51.0 \\

    VidEoMT~\cite{norouzi2026videomt} & ViT-L~\cite{dosovitskiy2021vit} & D2 & \twemoji{fire} &
    & 1897 & 75 & 55.2 & 48.9 \\

    VidEoMT~\cite{norouzi2026videomt} & ViT-L~\cite{dosovitskiy2021vit} & D3 & \twemoji{fire} &
    & 1897 & 71 & 55.1 & 48.1 \\

    \midrule

    CAVIS~\cite{lee2025cavis} & ViT-Adapter-L~\cite{chen2023vitadapter} & D2 & \twemoji{2744} &
    & 2612 & 10 & 56.4 & 49.0 \\

    \textbf{\oursShort} (Ours) & ViT-L~\cite{dosovitskiy2021vit} & D2 & \twemoji{2744} &
    & 2037 & 60 & 55.3 & 48.2 \\

    \midrule

    CAVIS~\cite{lee2025cavis} & ViT-Adapter-L~\cite{chen2023vitadapter} & D3 & \twemoji{2744} &
    & 2612 & 9 & 56.8 & 51.2 \\

    \textbf{\oursShort} (Ours) & ViT-L~\cite{dosovitskiy2021vit} & D3 & \twemoji{2744} &
    & 2037 & 58 & 55.5 & 49.2 \\

    \bottomrule
    \end{tabularx}
    \caption{\textbf{\oursShort{} for Online VPS on VIPSeg~\cite{kim2020vps}.} \textbf{PT}: Pre-training. \textbf{I}:
ImageNet-21K~\cite{deng2009imagenet}. \textbf{D2}: DINOv2~\cite{oquab2023dinov2}). \textbf{D3}: DINOv3~\cite{simeoni2025dinov3}.}
    \label{tab:sota_comparison_vipseg}
\end{table}

\begin{table}[t]
    \centering
    \scriptsize
    \renewcommand{\tabcolsep}{2pt}
    \begin{tabularx}{\linewidth}{llcc c cccc}
    \toprule
    \multirow{2}[2]{*}{Method} &
    \multirow{2}[2]{*}{Backbone} &
    \multirow{2}[2]{*}{PT} &
    \multirow{2}[2]{*}{Encoder} && 
    \multicolumn{4}{c}{VSPW \textit{val}~\cite{miao2021vspw}} \\
    \cmidrule{6-9}
    &&&&&
    GFLOPs & FPS & mVC\textsubscript{16} & mIoU \\
    \midrule

    DVIS~\cite{zhang2023dvis} & Swin-L~\cite{liu2021swin} & I & \twemoji{fire} &
    & 879 & 22 & 94.3 & 61.3 \\

    DVIS++~\cite{zhang2025dvis++} & ViT-Adapter-L~\cite{chen2023vitadapter} & D2 & \twemoji{fire} &
    & 2290 & 13 & 94.2 & 62.8 \\

    VidEoMT~\cite{norouzi2026videomt} & ViT-L~\cite{dosovitskiy2021vit} & D2 & \twemoji{fire} &
    & 1909 & 73 & 95.0 & 64.9 \\

    VidEoMT~\cite{norouzi2026videomt} & ViT-L~\cite{dosovitskiy2021vit} & D3 & \twemoji{fire} &
    & 1909 & 71 & 94.4 & 64.0 \\

    \midrule

    \textbf{\oursShort} (Ours) & ViT-L~\cite{dosovitskiy2021vit} & D2 & \twemoji{2744} &
    & 2049 & 60 & 94.6 & 64.3 \\

    \textbf{\oursShort} (Ours) & ViT-L~\cite{dosovitskiy2021vit} & D3 & \twemoji{2744} &
    & 2049 & 57 & 94.9 & 65.7 \\

    \bottomrule
    \end{tabularx}
    \caption{\textbf{\oursShort{} for Online VSS on VSPW~\cite{miao2021vspw}.} \textbf{PT}: Pre-training. \textbf{I}:
ImageNet-21K~\cite{deng2009imagenet}. \textbf{D2}: DINOv2~\cite{oquab2023dinov2}). \textbf{D3}: DINOv3~\cite{simeoni2025dinov3}.}
    \label{tab:sota_comparison_vspw}
\end{table}

\subsection{Video Segmentation Benchmarks}

\PARbegin{Video Instance Segmentation.} 
In \cref{tab:sota_comparison_ytvis19_ytvis21}, we compare \oursShort{} with state-of-the-art online VIS models on YouTube-VIS 2019 and 2021. 
Here, the baseline frozen-encoder model that we compare to is CAVIS~\cite{lee2025cavis}, a state-of-the-art model with many task-specific components that performs well when keeping the backbone frozen.
Compared to CAVIS, \oursShort{} obtains consistent speed gains of more than $8\times$, which are even larger than those observed for image segmentation, as video segmentation models use additional inefficient components.
For this task, \oursShort{} also consistently obtains a slightly higher accuracy than CAVIS when using a DINOv3 encoder, with 69.2 \vs 68.8 mAP for Youtube-VIS 2019.
Notably, \oursShort{} even performs better than the state-of-the-art CAVIS and VidEoMT models that \textit{finetune} the encoder, while being only slightly slower than VidEoMT. 
These results demonstrate the strength of \oursShort{}'s design for video segmentation, as they show that it can obtain state-of-the-art results at high inference speeds, while still allowing the frozen encoder to be reused for other tasks.

\PAR{Video Panoptic Segmentation.}
In \cref{tab:sota_comparison_vipseg}, we compare \oursShort{} with state-of-the-art VPS methods on VIPSeg.
As in the video instance segmentation setting, \oursShort{} is substantially faster than the frozen-encoder CAVIS model. 
In this case, \oursShort{}'s accuracy is slightly lower, with gaps of up to 1.3 VPQ, which we expect to be due to the higher complexity of the VPS task.
Again, however, \oursShort{} performs better than VidEoMT that finetunes the encoder, while only being slightly slower. 
These results offer further evidence of the effectiveness of \oursShort{} and its favorable accuracy \vs efficiency trade-off.

\PAR{Video Semantic Segmentation.}
In \cref{tab:sota_comparison_vspw}, we evaluate \oursShort{} for video semantic segmentation on VSPW.
Notably, \ours{} obtains new state-of-the-art results on this dataset in terms of the main mIoU metric, with 65.7 \vs the prior best of 64.9.
Impressively, this is obtained while keeping the encoder frozen.
This further demonstrates the strength of \oursShort{}'s design.

\section{Conclusion}
\label{sec:conclusion}
In this work, we reconcile the architectural simplicity and speed of encoder-only segmentation models with the practical need to keep Vision Foundation Model encoders frozen.
Our \decoder{} (\decoderShort{}) achieves this by replicating the joint query-patch attention of EoMT's last encoder blocks in a small module that operates entirely on top of frozen features.
The resulting model, \ours{} (\oursShort{}), applies seamlessly to both image and video segmentation: on standard image benchmarks it matches the frozen-encoder state of the art at up to ${\sim}3\times$ higher inference speed, while on video benchmarks it achieves accuracies on par with fully finetuned methods, at up to $8\times$ the speed of state-of-the-art frozen-encoder baselines.
Our ablations confirm that the effectiveness of the lightweight \decoderShort{} depends on a sufficiently large and extensively pre-trained encoder, in line with the scaling observations of EoMT~\cite{kerssies2025eomt} and VidEoMT~\cite{norouzi2026videomt}.
Because the encoder remains entirely untouched, \ours{} enables efficient multi-task deployment where a single shared frozen encoder serves multiple downstream tasks and datasets simultaneously.

\PAR{Acknowledgements.}
This work was partly funded by the Cynergy4MIE project, which is supported by the Chips Joint Undertaking and its members, including the top-up funding by National Authorities under Grant Agreement No 101140226.

{
    \small
    \bibliographystyle{ieeenat_fullname}
    \bibliography{main}

\begin{thebibliography}{53}
\providecommand{\natexlab}[1]{#1}
\providecommand{\url}[1]{\texttt{#1}}
\expandafter\ifx\csname urlstyle\endcsname\relax
  \providecommand{\doi}[1]{doi: #1}\else
  \providecommand{\doi}{doi: \begingroup \urlstyle{rm}\Url}\fi

\bibitem[Ansel et~al.(2024)Ansel, Yang, He, Gimelshein, Jain, Voznesensky, Bao, Bell, Berard, Burovski, Chauhan, Chourdia, Constable, Desmaison, DeVito, Ellison, Feng, Gong, Gschwind, Hirsh, Huang, Kalambarkar, Kirsch, Lazos, Lezcano, Liang, Liang, Lu, Luk, Maher, et~al.]{ansel2024pytorch2}
Jason Ansel, Edward Yang, Horace He, Natalia Gimelshein, Animesh Jain, Michael Voznesensky, Bin Bao, Peter Bell, David Berard, Evgeni Burovski, Geeta Chauhan, Anjali Chourdia, Will Constable, Alban Desmaison, Zachary DeVito, Elias Ellison, Will Feng, Jiong Gong, Michael Gschwind, Brian Hirsh, Sherlock Huang, Kshiteej Kalambarkar, Laurent Kirsch, Michael Lazos, Mario Lezcano, Yanbo Liang, Jason Liang, Yinghai Lu, C.~K. Luk, Bert Maher, et~al.
\newblock {PyTorch} 2: Faster machine learning through dynamic python bytecode transformation and graph compilation.
\newblock In \emph{ASPLOS}, 2024.

\bibitem[Ba et~al.(2016)Ba, Kiros, and Hinton]{ba2016layernorm}
Jimmy~Lei Ba, Jamie~Ryan Kiros, and Geoffrey~E. Hinton.
\newblock Layer normalization.
\newblock \emph{arXiv preprint arXiv:1607.06450}, 2016.

\bibitem[Bolya et~al.(2023)Bolya, Fu, Dai, Zhang, Feichtenhofer, and Hoffman]{bolya2023tome}
Daniel Bolya, Cheng-Yang Fu, Xiaoliang Dai, Peizhao Zhang, Christoph Feichtenhofer, and Judy Hoffman.
\newblock Token merging: Your {ViT} but faster.
\newblock In \emph{ICLR}, 2023.

\bibitem[Carion et~al.(2020)Carion, Massa, Synnaeve, Usunier, Kirillov, and Zagoruyko]{carion2020detr}
Nicolas Carion, Francisco Massa, Gabriel Synnaeve, Nicolas Usunier, Alexander Kirillov, and Sergey Zagoruyko.
\newblock End-to-end object detection with transformers.
\newblock In \emph{ECCV}, 2020.

\bibitem[Caron et~al.(2021)Caron, Touvron, Misra, J\'{e}gou, Mairal, Bojanowski, and Joulin]{caron2021dino}
Mathilde Caron, Hugo Touvron, Ishan Misra, Herv\'{e} J\'{e}gou, Julien Mairal, Piotr Bojanowski, and Armand Joulin.
\newblock Emerging properties in self-supervised vision transformers.
\newblock In \emph{ICCV}, 2021.

\bibitem[Cavagnero et~al.(2024)Cavagnero, Rosi, Cuttano, Pistilli, Ciccone, Averta, and Cermelli]{cavagnero2024pem}
Niccol\`{o} Cavagnero, Gabriele Rosi, Claudia Cuttano, Francesca Pistilli, Marco Ciccone, Giuseppe Averta, and Fabio Cermelli.
\newblock {PEM}: Prototype-based efficient {MaskFormer} for image segmentation.
\newblock In \emph{CVPR}, 2024.

\bibitem[Chen et~al.(2018{\natexlab{a}})Chen, Papandreou, Kokkinos, Murphy, and Yuille]{chen2018deeplab}
Liang-Chieh Chen, George Papandreou, Iasonas Kokkinos, Kevin Murphy, and Alan~L. Yuille.
\newblock {DeepLab}: Semantic image segmentation with deep convolutional nets, atrous convolution, and fully connected {CRFs}.
\newblock \emph{IEEE TPAMI}, 40\penalty0 (4):\penalty0 834--848, 2018{\natexlab{a}}.

\bibitem[Chen et~al.(2018{\natexlab{b}})Chen, Zhu, Papandreou, Schroff, and Adam]{chen2018deeplabv3p}
Liang-Chieh Chen, Yukun Zhu, George Papandreou, Florian Schroff, and Hartwig Adam.
\newblock Encoder-decoder with atrous separable convolution for semantic image segmentation.
\newblock In \emph{ECCV}, 2018{\natexlab{b}}.

\bibitem[Chen et~al.(2023)Chen, Duan, Wang, He, Lu, Dai, and Qiao]{chen2023vitadapter}
Zhe Chen, Yuchen Duan, Wenhai Wang, Junjun He, Tong Lu, Jifeng Dai, and Yu Qiao.
\newblock Vision transformer adapter for dense predictions.
\newblock In \emph{ICLR}, 2023.

\bibitem[Cheng et~al.(2021)Cheng, Schwing, and Kirillov]{cheng2021maskformer}
Bowen Cheng, Alex Schwing, and Alexander Kirillov.
\newblock Per-pixel classification is not all you need for semantic segmentation.
\newblock In \emph{NeurIPS}, 2021.

\bibitem[Cheng et~al.(2022)Cheng, Misra, Schwing, Kirillov, and Girdhar]{cheng2022mask2former}
Bowen Cheng, Ishan Misra, Alexander~G. Schwing, Alexander Kirillov, and Rohit Girdhar.
\newblock Masked-attention mask transformer for universal image segmentation.
\newblock In \emph{CVPR}, 2022.

\bibitem[Dao(2024)]{dao2024flashattention}
Tri Dao.
\newblock {FlashAttention-2}: Faster attention with better parallelism and work partitioning.
\newblock In \emph{ICLR}, 2024.

\bibitem[Deng et~al.(2009)Deng, Dong, Socher, Li, Li, and Fei-Fei]{deng2009imagenet}
Jia Deng, Wei Dong, Richard Socher, Li-Jia Li, Kai Li, and Li Fei-Fei.
\newblock {ImageNet}: A large-scale hierarchical image database.
\newblock In \emph{CVPR}, 2009.

\bibitem[Dosovitskiy et~al.(2021)Dosovitskiy, Beyer, Kolesnikov, Weissenborn, Zhai, Unterthiner, Dehghani, Minderer, Heigold, Gelly, Uszkoreit, and Houlsby]{dosovitskiy2021vit}
Alexey Dosovitskiy, Lucas Beyer, Alexander Kolesnikov, Dirk Weissenborn, Xiaohua Zhai, Thomas Unterthiner, Mostafa Dehghani, Matthias Minderer, Georg Heigold, Sylvain Gelly, Jakob Uszkoreit, and Neil Houlsby.
\newblock An image is worth 16x16 words: Transformers for image recognition at scale.
\newblock In \emph{ICLR}, 2021.

\bibitem[Hassani and Shi(2022)]{hassani2022dinat}
Ali Hassani and Humphrey Shi.
\newblock Dilated neighborhood attention transformer.
\newblock \emph{arXiv preprint arXiv:2209.15001}, 2022.

\bibitem[Heo et~al.(2023)Heo, Hwang, Hyun, Kim, Oh, Lee, and Kim]{GenVIS}
Miran Heo, Sukjun Hwang, Jeongseok Hyun, Hanjung Kim, Seoung~Wug Oh, Joon-Young Lee, and Seon~Joo Kim.
\newblock {A Generalized Framework for Video Instance Segmentation}.
\newblock In \emph{CVPR}, 2023.

\bibitem[Huang et~al.(2022)Huang, Yu, and Anandkumar]{huang2022minvis}
De-An Huang, Zhiding Yu, and Anima Anandkumar.
\newblock {MinVIS}: A minimal video instance segmentation framework without video-based training.
\newblock In \emph{NeurIPS}, 2022.

\bibitem[Ioffe and Szegedy(2015)]{ioffe2015batchnorm}
Sergey Ioffe and Christian Szegedy.
\newblock {Batch Normalization: Accelerating Deep Network Training by Reducing Internal Covariate Shift}.
\newblock 2015.

\bibitem[Jain et~al.(2023)Jain, Li, Chiu, Hassani, Orlov, and Shi]{jain2023oneformer}
Jitesh Jain, Jiachen Li, Mang~Tik Chiu, Ali Hassani, Nikita Orlov, and Humphrey Shi.
\newblock {OneFormer}: One transformer to rule universal image segmentation.
\newblock In \emph{CVPR}, 2023.

\bibitem[Kerssies et~al.(2025)Kerssies, Cavagnero, Hermans, Norouzi, Averta, Leibe, Dubbelman, and de~Geus]{kerssies2025eomt}
Tommie Kerssies, Niccol\`{o} Cavagnero, Alexander Hermans, Narges Norouzi, Giuseppe Averta, Bastian Leibe, Gijs Dubbelman, and Daan de Geus.
\newblock Your {ViT} is secretly an image segmentation model.
\newblock In \emph{CVPR}, 2025.

\bibitem[Kim et~al.(2020)Kim, Woo, Lee, and Kweon]{kim2020vps}
Dahun Kim, Sanghyun Woo, Joon-Young Lee, and In~So Kweon.
\newblock Video panoptic segmentation.
\newblock In \emph{CVPR}, 2020.

\bibitem[Kirillov et~al.(2019)Kirillov, He, Girshick, Rother, and Doll\'{a}r]{kirillov2019panoptic}
Alexander Kirillov, Kaiming He, Ross Girshick, Carsten Rother, and Piotr Doll\'{a}r.
\newblock Panoptic segmentation.
\newblock In \emph{CVPR}, 2019.

\bibitem[Lee et~al.(2025)Lee, Seo, Han, Choi, and Im]{lee2025cavis}
Seunghun Lee, Jiwan Seo, Kiljoon Han, Minwoo Choi, and Sunghoon Im.
\newblock Context-aware video instance segmentation.
\newblock In \emph{ICCV}, 2025.

\bibitem[Li et~al.(2023)Li, Zhang, Xu, Liu, Zhang, Ni, and Shum]{li2023maskdino}
Feng Li, Hao Zhang, Huaizhe Xu, Shilong Liu, Lei Zhang, Lionel~M. Ni, and Heung-Yeung Shum.
\newblock {Mask DINO}: Towards a unified transformer-based framework for object detection and segmentation.
\newblock In \emph{CVPR}, 2023.

\bibitem[Lin et~al.(2014)Lin, Maire, Belongie, Hays, Perona, Ramanan, Doll\'{a}r, and Zitnick]{lin2014coco}
Tsung-Yi Lin, Michael Maire, Serge Belongie, James Hays, Pietro Perona, Deva Ramanan, Piotr Doll\'{a}r, and C.~Lawrence Zitnick.
\newblock Microsoft {COCO}: Common objects in context.
\newblock In \emph{ECCV}, 2014.

\bibitem[Liu et~al.(2021)Liu, Lin, Cao, Hu, Wei, Zhang, Lin, and Guo]{liu2021swin}
Ze Liu, Yutong Lin, Yue Cao, Han Hu, Yixuan Wei, Zheng Zhang, Stephen Lin, and Baining Guo.
\newblock Swin transformer: Hierarchical vision transformer using shifted windows.
\newblock In \emph{ICCV}, 2021.

\bibitem[Liu et~al.(2022)Liu, Mao, Wu, Feichtenhofer, Darrell, and Xie]{liu2022convnext}
Zhuang Liu, Hanzi Mao, Chao-Yuan Wu, Christoph Feichtenhofer, Trevor Darrell, and Saining Xie.
\newblock A {ConvNet} for the 2020s.
\newblock In \emph{CVPR}, 2022.

\bibitem[Long et~al.(2015)Long, Shelhamer, and Darrell]{long2015fcn}
Jonathan Long, Evan Shelhamer, and Trevor Darrell.
\newblock Fully convolutional networks for semantic segmentation.
\newblock In \emph{CVPR}, 2015.

\bibitem[Loshchilov and Hutter(2017)]{loshchilov2017sgdr}
Ilya Loshchilov and Frank Hutter.
\newblock {SGDR: Stochastic Gradient Descent with Warm Restarts}.
\newblock In \emph{ICLR}, 2017.

\bibitem[Loshchilov and Hutter(2019)]{loshchilov2019adamw}
Ilya Loshchilov and Frank Hutter.
\newblock Decoupled weight decay regularization.
\newblock In \emph{ICLR}, 2019.

\bibitem[Lu et~al.(2023)Lu, de~Geus, and Dubbelman]{lu2023cts}
Chenyang Lu, Daan de Geus, and Gijs Dubbelman.
\newblock {Content-Aware Token Sharing for Efficient Semantic Segmentation With Vision Transformers}.
\newblock In \emph{CVPR}, 2023.

\bibitem[{Meta Research}(2023)]{metaresearch2023fvcore}
{Meta Research}.
\newblock fvcore, 2023.

\bibitem[Miao et~al.(2021)Miao, Wei, Wu, Liang, Li, and Yang]{miao2021vspw}
Jiaxu Miao, Yunchao Wei, Yu Wu, Chen Liang, Guangrui Li, and Yi Yang.
\newblock {VSPW: A Large-scale Dataset for Video Scene Parsing in the Wild}.
\newblock In \emph{CVPR}, 2021.

\bibitem[Miao et~al.(2022)Miao, Wang, Wu, Li, Zhang, Wei, and Yang]{miao2022large}
Jiaxu Miao, Xiaohan Wang, Yu Wu, Wei Li, Xu Zhang, Yunchao Wei, and Yi Yang.
\newblock {Large-Scale Video Panoptic Segmentation in the Wild: A Benchmark}.
\newblock In \emph{CVPR}, 2022.

\bibitem[Milletari et~al.(2016)Milletari, Navab, and Ahmadi]{milletari2016dice}
Fausto Milletari, Nassir Navab, and Seyed-Ahmad Ahmadi.
\newblock {V-Net}: Fully convolutional neural networks for volumetric medical image segmentation.
\newblock In \emph{3DV}, 2016.

\bibitem[Nilsson and Sminchisescu(2018)]{nilsson2018vss}
David Nilsson and Cristian Sminchisescu.
\newblock Semantic video segmentation by gated recurrent flow propagation.
\newblock In \emph{CVPR}, 2018.

\bibitem[Norouzi et~al.(2024)Norouzi, Orlova, de~Geus, and Dubbelman]{norouzi2024algm}
Narges Norouzi, Svetlana Orlova, Daan de Geus, and Gijs Dubbelman.
\newblock {ALGM}: Adaptive local-then-global token merging for efficient semantic segmentation with plain vision transformers.
\newblock In \emph{CVPR}, 2024.

\bibitem[Norouzi et~al.(2026)Norouzi, Zulfikar, Cavagnero, Kerssies, Leibe, Dubbelman, and de~Geus]{norouzi2026videomt}
Narges Norouzi, Idil~Esen Zulfikar, Niccol\`{o} Cavagnero, Tommie Kerssies, Bastian Leibe, Gijs Dubbelman, and Daan de Geus.
\newblock {VidEoMT}: Your {ViT} is secretly also a video segmentation model.
\newblock In \emph{CVPR}, 2026.

\bibitem[Oquab et~al.(2024)Oquab, Darcet, Moutakanni, Vo, Szafraniec, Khalidov, Fernandez, Haziza, Massa, El-Nouby, et~al.]{oquab2023dinov2}
Maxime Oquab, Timoth\'{e}e Darcet, Th\'{e}o Moutakanni, Huy Vo, Marc Szafraniec, Vasil Khalidov, Pierre Fernandez, Daniel Haziza, Francisco Massa, Alaaeldin El-Nouby, et~al.
\newblock {DINOv2}: Learning robust visual features without supervision.
\newblock \emph{TMLR}, 2024.

\bibitem[Russakovsky et~al.(2015)Russakovsky, Deng, Su, Krause, Satheesh, Ma, Huang, Karpathy, Khosla, Bernstein, et~al.]{russakovsky2015imagenet}
Olga Russakovsky, Jia Deng, Hao Su, Jonathan Krause, Sanjeev Satheesh, Sean Ma, Zhiheng Huang, Andrej Karpathy, Aditya Khosla, Michael Bernstein, et~al.
\newblock {ImageNet} large scale visual recognition challenge.
\newblock \emph{IJCV}, 2015.

\bibitem[Sim\'{e}oni et~al.(2025)Sim\'{e}oni, Vo, Seitzer, Baldassarre, Oquab, Jose, Khalidov, Szafraniec, Yi, Ramamonjisoa, Massa, Haziza, Wehrstedt, Wang, Darcet, Moutakanni, Sentana, Roberts, Vedaldi, Tolan, Brandt, Couprie, Mairal, J\'{e}gou, Labatut, and Bojanowski]{simeoni2025dinov3}
Oriane Sim\'{e}oni, Huy~V. Vo, Moritz Seitzer, Filippo Baldassarre, Maxime Oquab, Camille Jose, Vasil Khalidov, Marc Szafraniec, Souyoung Yi, Manu Ramamonjisoa, Francisco Massa, Daniel Haziza, Loubna Wehrstedt, Jianyuan Wang, Timoth\'{e}e Darcet, Th\'{e}o Moutakanni, Lorena Sentana, Cameron Roberts, Andrea Vedaldi, James Tolan, Jasper Brandt, Camille Couprie, Julien Mairal, Herv\'{e} J\'{e}gou, Pierre Labatut, and Piotr Bojanowski.
\newblock {DINOv3}.
\newblock \emph{arXiv preprint arXiv:2508.10104}, 2025.

\bibitem[Su et~al.(2024)Su, Ahmed, Lu, Pan, Bo, and Liu]{su2024roformer}
Jianlin Su, Murtadha Ahmed, Yu Lu, Shengfeng Pan, Wen Bo, and Yunfeng Liu.
\newblock {RoFormer: Enhanced Transformer with Rotary Position Embedding}.
\newblock \emph{Neurocomputing}, 568, 2024.

\bibitem[Vaswani et~al.(2017)Vaswani, Shazeer, Parmar, Uszkoreit, Jones, Gomez, Kaiser, and Polosukhin]{vaswani2017attention}
Ashish Vaswani, Noam Shazeer, Niki Parmar, Jakob Uszkoreit, Llion Jones, Aidan~N. Gomez, {\L}ukasz Kaiser, and Illia Polosukhin.
\newblock Attention is all you need.
\newblock In \emph{NeurIPS}, 2017.

\bibitem[Wang et~al.(2021)Wang, Zhu, Adam, Yuille, and Chen]{wang2021maxdeeplab}
Huiyu Wang, Yukun Zhu, Hartwig Adam, Alan Yuille, and Liang-Chieh Chen.
\newblock {MaX-DeepLab}: End-to-end panoptic segmentation with mask transformers.
\newblock In \emph{CVPR}, 2021.

\bibitem[Weber et~al.(2021)Weber, Xie, Collins, Zhu, Voigtlaender, Adam, Green, Geiger, Leibe, Cremers, et~al.]{weber2021step}
Mark Weber, Jun Xie, Maxwell Collins, Yukun Zhu, Paul Voigtlaender, Hartwig Adam, Bradley Green, Andreas Geiger, Bastian Leibe, Daniel Cremers, et~al.
\newblock {STEP: Segmenting and Tracking Every Pixel}.
\newblock 2021.

\bibitem[Xia et~al.(2024)Xia, Wang, Lv, Hao, and Shi]{xia2024vitcomer}
Chunlong Xia, Xinliang Wang, Feng Lv, Xin Hao, and Yifeng Shi.
\newblock {ViT-CoMer}: Vision transformer with convolutional multi-scale feature interaction for dense predictions.
\newblock In \emph{CVPR}, 2024.

\bibitem[Yang et~al.(2019{\natexlab{a}})Yang, Fan, and Xu]{yang2019video}
Linjie Yang, Yuchen Fan, and Ning Xu.
\newblock {Video Instance Segmentation}.
\newblock In \emph{ICCV}, 2019{\natexlab{a}}.

\bibitem[Yang et~al.(2019{\natexlab{b}})Yang, Fan, and Xu]{yang2019vis}
Linjie Yang, Yuchen Fan, and Ning Xu.
\newblock Video instance segmentation.
\newblock In \emph{ICCV}, 2019{\natexlab{b}}.

\bibitem[Yu et~al.(2022)Yu, Wang, Qiao, Collins, Zhu, Adam, Yuille, and Chen]{yu2022kmaxdeeplab}
Qihang Yu, Huiyu Wang, Siyuan Qiao, Maxwell Collins, Yukun Zhu, Hartwig Adam, Alan Yuille, and Liang-Chieh Chen.
\newblock k-means mask transformer.
\newblock In \emph{ECCV}, 2022.

\bibitem[Zhang et~al.(2023)Zhang, Tian, Wu, Ji, Wang, Zhang, and Wan]{zhang2023dvis}
Tao Zhang, Xingye Tian, Yu Wu, Shunping Ji, Xuebo Wang, Yuan Zhang, and Pengfei Wan.
\newblock {DVIS}: Decoupled video instance segmentation framework.
\newblock In \emph{CVPR}, 2023.

\bibitem[Zhang et~al.(2025)Zhang, Tian, Zhou, Ji, Wang, Tao, Zhang, Wan, Wang, and Wu]{zhang2025dvis++}
Tao Zhang, Xingye Tian, Yikang Zhou, Shunping Ji, Xuebo Wang, Xin Tao, Yuan Zhang, Pengfei Wan, Zhongyuan Wang, and Yu Wu.
\newblock {DVIS++: Improved Decoupled Framework for Universal Video Segmentation}.
\newblock \emph{PAMI}, 2025.

\bibitem[Zhou et~al.(2017)Zhou, Zhao, Puig, Fidler, Barriuso, and Torralba]{zhou2017ade20k}
Bolei Zhou, Hang Zhao, Xavier Puig, Sanja Fidler, Adela Barriuso, and Antonio Torralba.
\newblock Scene parsing through {ADE20K} dataset.
\newblock In \emph{CVPR}, 2017.

\bibitem[Zhou et~al.(2024)Zhou, Zhang, Ji, Yan, and Li]{zhou2024dvisdaq}
Yikang Zhou, Tao Zhang, Shunping Ji, Shuicheng Yan, and Xiangtai Li.
\newblock Improving video segmentation via dynamic anchor queries.
\newblock In \emph{ECCV}, 2024.

\end{thebibliography}
}

\end{document}